\documentclass[10pt]{article}

\usepackage{times}
\usepackage{amsmath}
\usepackage{amsfonts}
\usepackage{amssymb}
\usepackage[hyphens]{url}
\usepackage{latexsym}
\usepackage{color}
\usepackage{graphicx} 
\usepackage{subfigure} 
\usepackage{natbib}
\usepackage{algorithm}
\usepackage{algorithmic}

\usepackage{hyperref}

\usepackage[accepted]{icml2015} 

\icmltitlerunning{BilBOWA: Fast Bilingual Distributed Representations without Word Alignments}

\newcommand\plpl{\texttt{++}}

\newcommand\bld[1]{\textbf{#1}}
\newcommand\mat[1]{\mathbf{#1}}

\newcommand\loss{\mathcal{L}}

\DeclareMathOperator*{\E}{\mathbb{E}}

\begin{document} 

\twocolumn[
\icmltitle{BilBOWA: Fast Bilingual Distributed Representations without Word Alignments}

\icmlauthor{Stephan Gouws}{sgouws@google.com}
\icmladdress{Google Inc., Mountain View, CA, USA}
\icmlauthor{Yoshua Bengio}{}
\icmladdress{Dept.\ IRO, Universit\'e de Montr\'eal, QC, Canada \& Canadian Institute for Advanced Research}
\icmlauthor{Greg Corrado}{}
\icmladdress{Google Inc., Mountain View, CA, USA}

\icmlkeywords{word embedding, translation, neural language model, word2vec}

\vskip 0.3in
]

\begin{abstract} 
We introduce BilBOWA (\emph{Bilingual Bag-of-Words without Alignments}), a
simple and computationally-efficient model for learning bilingual distributed
representations of words which can scale to large monolingual datasets and does not require
word-aligned parallel training data. Instead it trains directly on monolingual data
and extracts a bilingual signal from a smaller set of raw-text sentence-aligned
data. This is achieved using a novel sampled bag-of-words cross-lingual
objective, which is used to regularize two noise-contrastive language models
for efficient cross-lingual feature learning. We show that bilingual embeddings learned
using the proposed model outperform state-of-the-art methods on a
cross-lingual document classification task as well as a lexical translation
task on WMT11 data.
\end{abstract} 

\section{Introduction}
\label{sec:intro}

\begin{figure*}
  \begin{center}
    \includegraphics[width=0.9\textwidth]{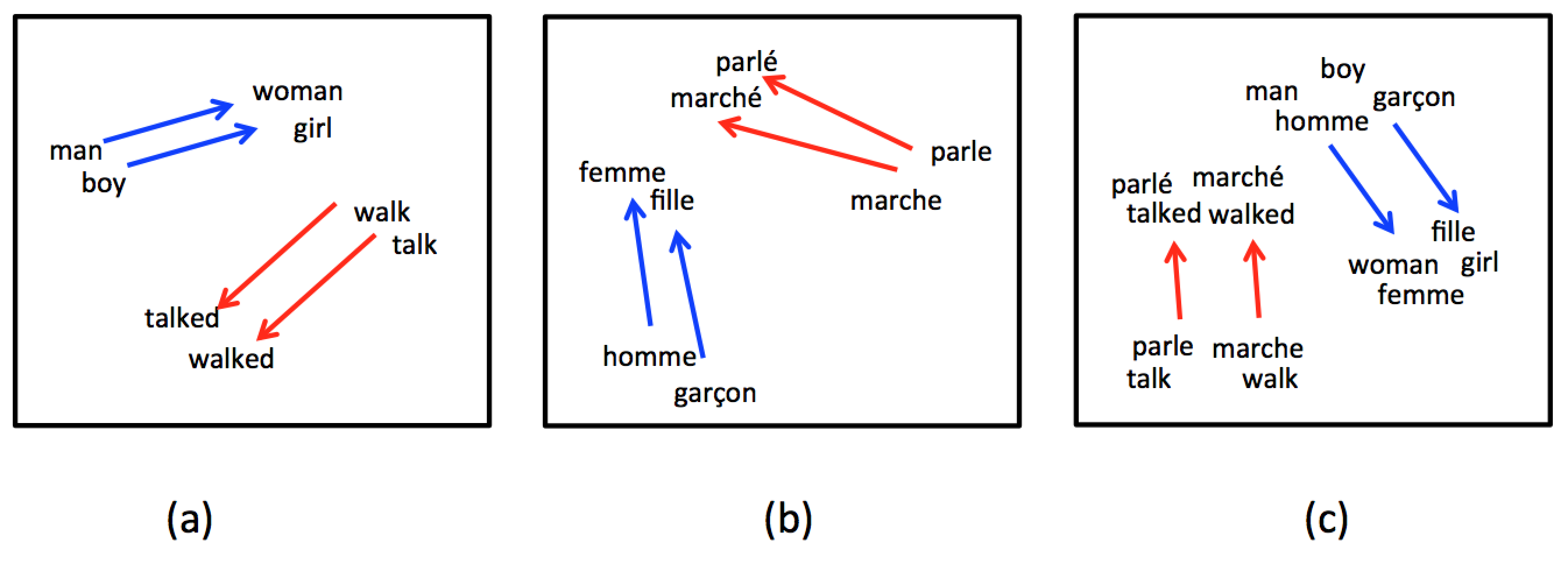}
  \end{center}
  \caption{(a \& b) Monolingual embeddings have been shown to capture syntactic
  and semantic features such as noun gender (blue) and verb tense
  (red).  (c) The (idealized) goal of crosslingual embeddings is to capture these
  relationships across two or more languages.}
  \label{fig:crossling-embeddings}
\end{figure*}

Raw text data is freely available in many languages, yet labeled data --
e.g.\ text marked up with parts-of-speech or named-entities -- is
expensive and mostly available for English. Although several techniques exist
that can learn to map hand-crafted features from one domain to another
\cite{blitzer2006domain,daume2009frustratingly,pan2010survey}, it is in general
non-trivial to come up with good features which generalize well across tasks,
and even harder across different languages. It is therefore very desirable to
have unsupervised techniques which can learn useful syntactic and semantic
features that are invariant to the tasks or languages that we are interested
in.  Unsupervised \emph{distributed} representations of words capture important
syntactic and semantic information about languages and these techniques have
been succesfully applied to a wide range of tasks
\cite{collobert2011natural,turian2010word}, across many different languages
\cite{polyglot2013}. Traditionally, inducing these representations involved
training a neural network language model \cite{bengio2003neural} which was slow
to train. However, contemporary word embedding models are much faster in comparison,
and can scale to train on billions of words per day on a single desktop machine
~\cite{mnih2012fast,mikolov2013distributed,pennington2014glove}. In
all these models, words are represented by learned, real-valued feature vectors
referred to as \emph{word embeddings} and trained from large amounts of raw text.
These models have the property that similar embedding vectors are learned for
similar words during training. Additionally, the vectors capture rich linguistic relationships
such as \emph{male-female} relationships or verb tenses, as illustrated in 
Figure~\ref{fig:crossling-embeddings}~(a) and (b).  These two properties improve 
generalization when the embedding
vectors are used as features on word- and sentence-level prediction tasks.

Distributed representations can also be induced \emph{over different
language-pairs} and can serve as an effective way of learning linguistic
regularities which generalize across languages, in that words with similar
distributional syntactic and semantic properties in both languages are
represented using similar vectorial representations (i.e.\ embed nearby in the
embedded space, as shown in Figure~\ref{fig:crossling-embeddings}~(c)).  This
is especially useful for transferring limited label information from
high-resource to low-resource languages, and has been demonstrated to be
effective for document classification \cite{klementiev2012}, outperforming a
strong machine-translation baseline; as well as named-entity recognition and
machine translation \cite{zoubilingual,mikolov2013exploiting}.  

Since these techniques are fundamentally data-driven techniques, the quality of
the learned representations improves as the size of the training data
improves~\cite{mikolov2013distributed,pennington2014glove}.  However, as we
will discuss in more detail in \S \ref{sec:related-work}, there are two
significant drawbacks associated with current bilingual embedding methods: they
are either very slow to train or they can only exploit parallel training data.
The former limits the large-scale application of these techniques, while the
latter severely limits the amount of available training data, and
furthermore introduces a big domain bias into the learning process, since parallel
data is typically only easily available for certain narrow domains (such as
parliamentary discussions).


This paper introduces \textbf{BilBOWA} (\emph{Bilingual
Bag-of-Words without Word Alignments}), a simple, scalable
technique for inducing bilingual word embeddings with a trivial extension to
multilingual embeddings. The model is able to leverage essentially unlimited 
amounts of monolingual raw text. It furthermore does not require any word-level 
alignments, but instead extracts a bilingual signal directly from a limited sample of
sentence-aligned, raw-text parallel data (e.g.\ Europarl) which it uses to
align embeddings as they are learned over monolingual training data. Our
contributions are the following:

\begin{itemize}
  \item We introduce a novel, computationally-efficient \emph{sampled
    cross-lingual objective} (``BilBOWA-loss'') which is employed to
    align monolingual embeddings as they are being trained in an online
    setting. The monolingual models can scale to large-scale training sets,
    thereby avoiding training bias, and the BilBOWA-loss only
    considers sampled bag-of-words sentence-aligned data at each training step,
    which scales extremely well and also avoids the need for estimating
    word-alignments (\S \ref{sec:bilbowa-xling});

  \item we experimentally evaluate the induced cross-lingual embeddings on a
    document-classification (\S \ref{sec:cldc}) and lexical translation task 
    (\S \ref{sec:wmt}), where the method outperforms current 
    state-of-the-art methods, with training time reduced to minutes or hours 
    compared to several days for prior approaches;

  \item finally, we make available our efficient C-implementation\footnote{
    \url{https://github.com/gouwsmeister/bilbowa}} to hopefully stimulate 
    further research on cross-lingual distributed feature learning.
\end{itemize}

\section{Learning Cross-lingual Word Embeddings}
\label{sec:related-work}

Monolingual word embedding algorithms
\cite{mikolov2013distributed,pennington2014glove} learn useful features about
words from raw text (e.g.\ Fig \ref{fig:crossling-embeddings} (a) \& (b)).  These
algorithms are trained over large datasets to be able to predict words from the
contexts in which they appear. Their working can
intuitively be understood as mapping each word to a learned vector in an
embedded space, and updating these vectors in an attempt to simultaneously
minimize the distance from a word's vector to the vectors of the words with
which it frequently co-occurs. The result of this optimization process yields 
a rich geometrical encoding of the distributional properties of natural language, 
where words with similar distributional properties cluster together. Due to 
their general nature, these features work well for several NLP prediction tasks 
\cite{collobert2011natural,turian2010word}.

In the cross-lingual setup, the goal is to learn features which generalize well
\emph{across different tasks and different languages}.  The goal is therefore
to learn features (embeddings) for each word 
such that \emph{similar words in each language} are assigned similar embeddings
(the \textbf{monolingual objectives}), but additionally we also want
\emph{similar words across languages} to have similar representations (the
\textbf{cross-lingual objective}). The latter property allows one to use the
learned embeddings as features for training a discriminative classifier to
predict labels in one language (e.g.\ topics, parts-of-speech, or
named-entities) where we have labelled data, and then directly transfer it to a
language for which we do not have much labelled data. From an optimization
perspective, there are several approaches to how one can optimize these two
objectives (our classification):

\textsc{\bld{Offline Alignment:}} The simplest approach is to optimize each
monolingual objective separately (i.e.\ train embeddings on each language
separately using any of the several available off-the-shelve toolkits), 
and then enforce the cross-lingual constraints as a separate,
disjoint, `alignment' step. The alignment step consists of learning a
transformation for projecting the embeddings of words onto the embeddings of their
translation pairs, obtained from a dictionary.  This was shown to be a viable
approach by \cite{mikolov2013exploiting} who learned a linear
projection from one embedding space to the other. It was extended by 
\cite{faruqui2014improving}, who simultanteously projected source and
target language embeddings into a joint space using canonical correlation
analysis. The advantage of this approach is that it is very fast to learn the
embedding alignments. The main drawback of this approach is that it is not
clear that a single transformation (whether linear or nonlinear) can capture
the relationships between all words in the source and target languages, and 
our improved results on the translation task seem to point to the contrary (\S \ref{sec:wmt}).
Furthermore, an accurate dictionary is required for the language-pair and the
method considers only one translation per word, which ignores the rich multi-sense
polysemy of natural languages.

\textsc{\bld{Parallel-only:}} Alternatively, one may leverage purely sentence-aligned
parallel data and train a model to learn similar representations for the
aligned sentences. This is the approach followed by the BiCVM \cite{hermann2013multilingual}
and the bilingual auto-encoder (BAE, \cite{chandar2014autoencoder}). 
The advantage of this approach is that it can be fast when using an efficient 
noise-contrastive training criterion like that of the BiCVM. The main drawbacks of
this method are that it can only train on \emph{limited parallel data}, which is
expensive to obtain and not necessarily written in the same style or register
as the domain where the features might be applied (i.e.\ there is a strong
\emph{domain bias}).

\begin{figure}
    \centering{
    \includegraphics[scale=0.25]{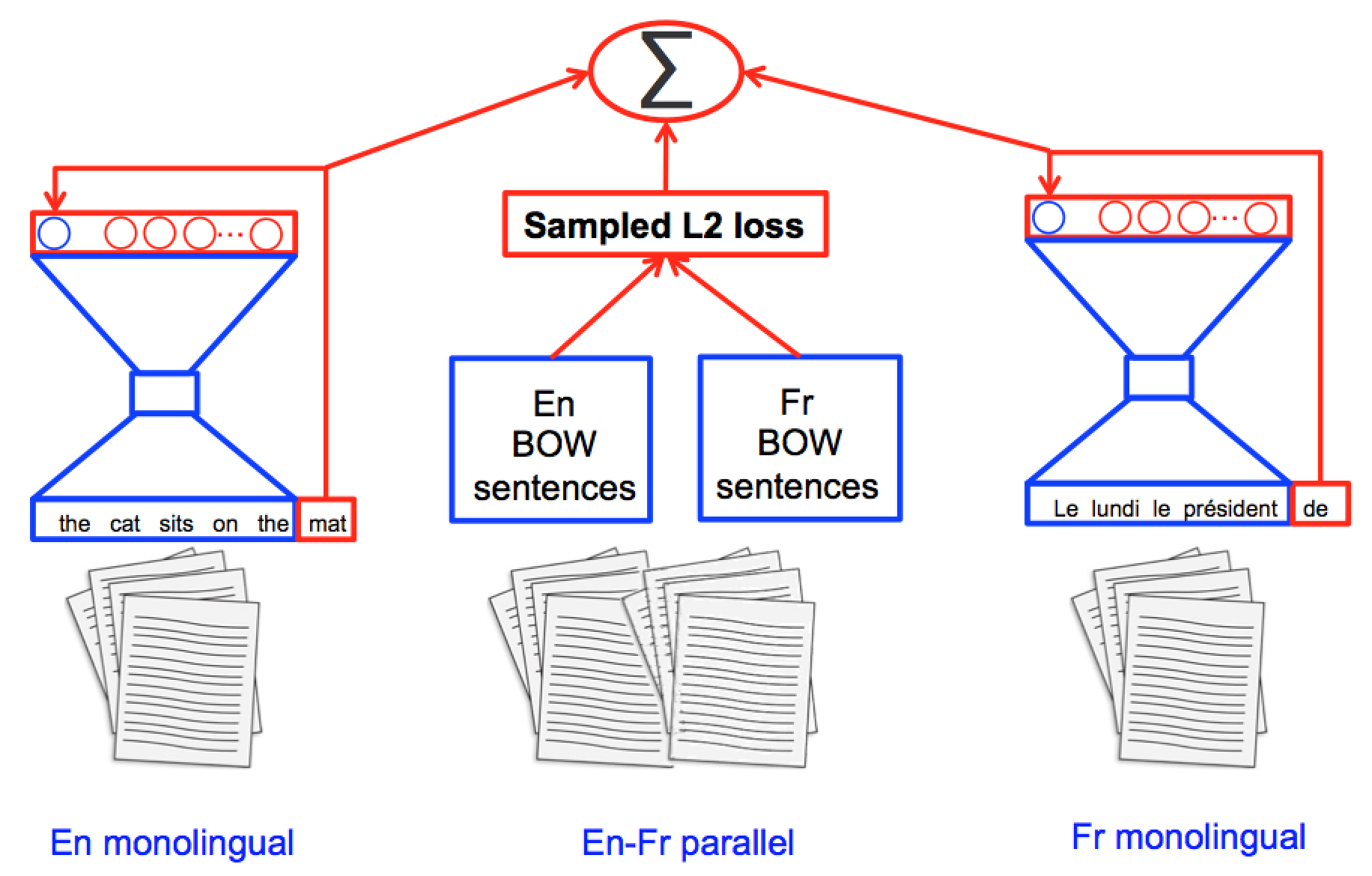}}
    \caption{Schematic of the proposed BilBOWA model architecture for inducing 
    bilingual word embeddings. Two monolingual skipgram models are jointly trained
    while enforcing a sampled $L_2$-loss which aligns the embeddings such that   
    translation-pairs are assigned similar embeddings in the two languages.}
\label{fig:bilbowa-arch}
\end{figure}

\textsc{\bld{Jointly-trained Models:}} Another approach is to \emph{jointly optimize}
the monolingual objectives $\loss(\cdot)$, with the cross-lingual objective
enforced as a \bld{cross-lingual regularizer} (see
Figure~\ref{fig:bilbowa-arch} for a schematic).  To do this, we define a
cross-lingual regularization term $\Omega(\cdot)$, and use it to constrain
monolingual models as they are jointly being trained over the context $h$
and target word $w_t$ training pairs in the dataset
$\mathcal{D}$, e.g.\: 
\begin{equation}
  \label{eqn:xling-general}
  \loss = \min_{\theta^e,\theta^f} \sum_{l \in \{e,f\}} \sum_{w_t,h \in \mathcal{D}^l} 
   \underbrace{\loss^l(w_t, h; \theta^l)}_\text{feature learning} + 
    \lambda \underbrace{\Omega(\theta^e, \theta^f)}_\text{alignment}.
\end{equation}

This formulation captures the intuition that we
want to learn representations which model their individual languages well (the
first term) while the $\Omega(\cdot)$ regularizer encourages representations to
be similar for words that are related across the two languages.  Conceptually,
this regularizer consists of minimizing a distance function between the
vector representations $\mat{r}_i$ learned for words $w_i$ in the two domains, weighted
by how similar they are, i.e.\ 
\begin{equation}
  \Omega(\mat{R}^e, \mat{R}^f) = 
  \sum_{w_i \in V^e} \sum_{w_j \in V^f} \textrm{sim}(w_i,w_j) 
     \cdot \text{distance}(\mat{r}^e_i, \mat{r}^f_j).
\end{equation}
where we use $\mat{R}$ to denote learned embedding representations, and $\mat{r}_i$ to denote
the embedding learned for word $w_i$. In other words, when this weighted sum (and
hence its contribution to the total objective) is low, one can be sure that
words across languages that are similar (i.e.\ high $\textrm{sim}(w_i, w_j)$) will
be embedded nearby each other. 

This approach was shown to be useful by \cite{klementiev2012}.  
Crucially, the advantages of this formulation are that it
\emph{enables one to train on any available monolingual data}, which is both more
abundant and less biased than the parallel-only approach, since one can train
on data which resembles the data one will be applying the learned features to.
The disadvantage is that the original model of Klementiev et al.\ is extremely
slow to train. The training complexity stems both from how the authors implement their
monolingual and cross-lingual objectives. For the monolingual objective, they
train a standard neural language model for which the complexity of the output 
softmax layer grows with the output vocabulary size. Therefore, in order to evaluate the model
the authors had to reduce the output vocabulary to only the $3000$ most frequent
words. The second reason for the slow training time is that the
cross-lingual objective considers the interactions between all pairs of words
(in the worst case) between the source and target vocabulary \emph{at each training step}, 
which scales as the product of the two vocabularies\footnote{If we 
limit each word to align to $k$ words this is still $O(Vk)$.}. In this work, we address these two issues
individually.

\section{The BilBOWA Model}
\label{sec:bilbowa}

As discussed in \S \ref{sec:related-work}, the primary challenges with existing
bilingual embedding models are their \emph{computational complexity} (due to an
expensive softmax or an expensive regularization term, or both), but most
importantly, the strong \emph{domain bias} that is introduced by models that
train only on parallel data such as Europarl. 

The BilBOWA model is designed to overcome these issues in order to enable 
computationally-efficient cross-lingual distributed feature learning over large 
amounts of monolingual text. A schematic overview of the
model is shown in Figure~\ref{fig:bilbowa-arch}. The two main aspects (discussed in the
following sections) are
\begin{enumerate}
  \item First, similar to \cite{zoubilingual}, we leverage advances in
    monolingual feature learning algorithms by replacing the softmax objective
    with a more efficient noise-contrastive objective (\S
    \ref{sec:bilbowa-monoling}), allowing monolingual training updates to scale
    independently of the vocabulary size.
  \item Second, we introduce a novel computationally-efficient cross-lingual loss 
    which only considers sampled, bag-of-words sentence-aligned data
    for the cross-lingual objective (\S \ref{sec:bilbowa-xling}). This avoids
    the need for estimating word alignments, but moreover, the computation of
    the regularization term reduces to only the words in the observed sample
    (compared to considering the $O(V^2)$ worst-case possible interactions at
    each training step in the naive case). 
\end{enumerate}

\subsection{Learning Monolingual Features: The $\loss$ term}
\label{sec:bilbowa-monoling}

Since we do not care about language modelling, but more about feature learning,
an alternative to the softmax is to use a \bld{noise-contrastive approach} to
score valid, observed combinations of words against randomly sampled, unlikely
combinations of words. This idea was introduced by Collobert and
Weston~\cite{collobert2011natural} where they optimized a margin between the
observed score and the noise scores. In their formulation, scores were computed
on \emph{sequences} of words, but in \cite{mikolov2013exploiting} this idea was taken one step further and
successfully applied to \emph{bag-of-word} representations of contexts in their
continuous bag-of-words (CBOW) and skipgram models trained using the negative 
sampling training objective (a simplified version of noise-contrastive estimation~\cite{mnih2012fast}). 
Any of these
objectives would yield comparable speedup and could be used in our architecture. In
this work we opted for the skipgram model trained using negative sampling since it
has been shown to learn high-quality monolingual features.

\subsection{Learning Cross-lingual Features: The BilBOWA-loss ($\Omega$ term)}
\label{sec:bilbowa-xling}

\begin{figure*}
  \begin{center}
    \includegraphics[width=0.8\textwidth]{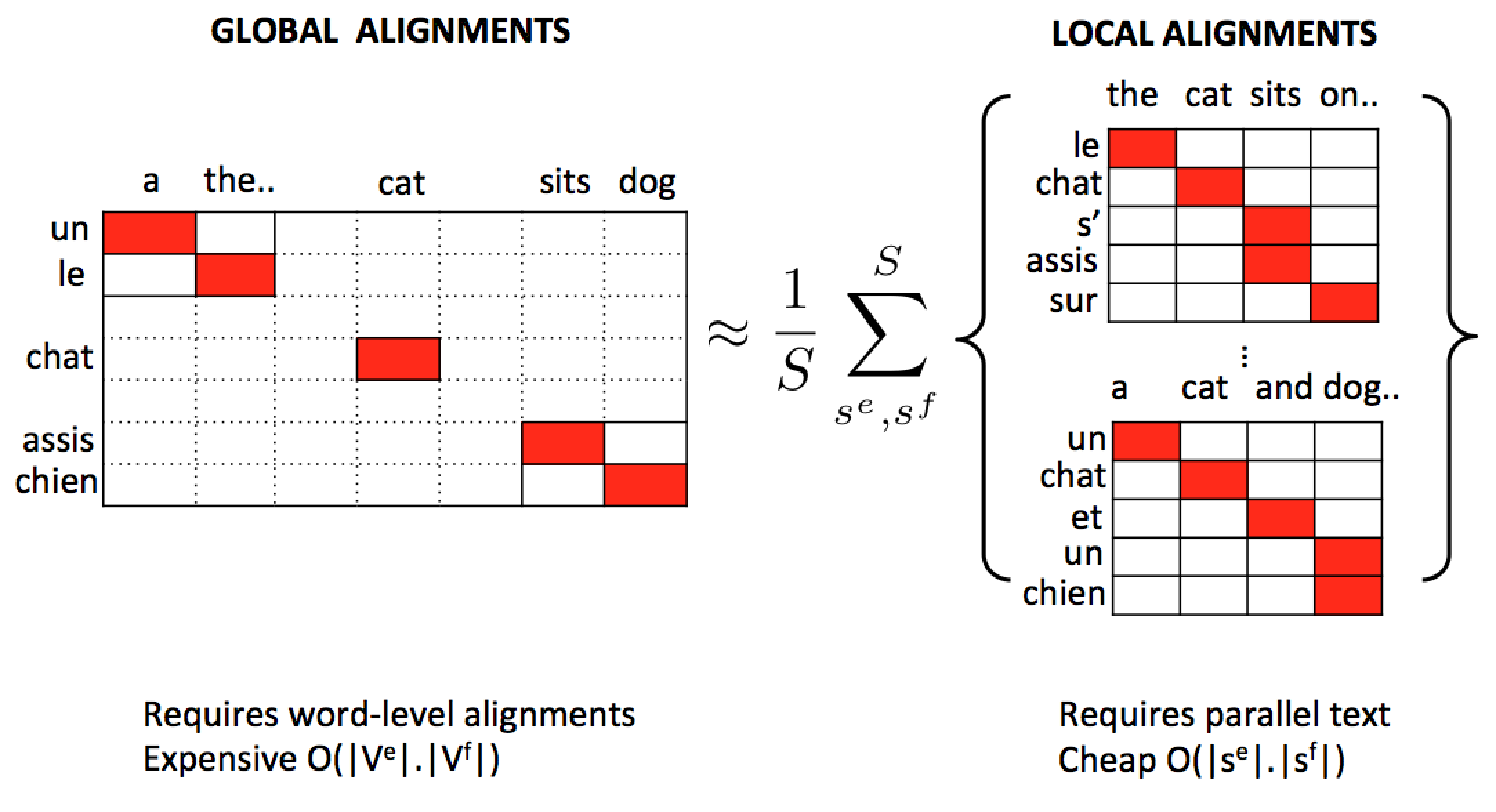}
  \end{center}
  \caption{Using global word-alignments for aligning cross-lingual embeddings 
  (equation \ref{eqn:xling-full}) is costly 
  and scales as the product of the two vocabulary sizes. In contrast, the BilBOWA-loss 
  (equation \ref{eqn:bilbowa-xling-sentvec}) approximates the global loss by averaging over 
  implicit local co-occurrence statistics in a limited sample of parallel sentence-pairs.}
  \label{fig:bilbowa-approximation}
\end{figure*}

Besides learning how words in one language relate to each other, it is equally 
important for the representations to capture how words between the two languages
relate to each other, which we enforce using the $\Omega$ term in equation
\ref{eqn:xling-general}.  In the general bilingual setting, word
similarities can be expressed as a matrix $\mat{A}$ where $a_{ij}$ encodes the
translation ``score'' of word $i$ in one language with word $j$ in the other.
In the rest of our discussion we will refer to English and French, and denote all 
English-specific parameters using \emph{e} superscript, and all French-specific parameters
with \emph{f} superscript. 

If the $K$-dimensional word embedding row-vectors $\mat{r}_i$
are stacked to form a $(V,K)$-dimensional matrix $\mat{R}$, then we can express
what we will refer to as the \bld{exact cross-lingual objective} as follows: 
\begin{align} \label{eqn:xling-full}
  \Omega_\mat{A}(\mat{R}^e, \mat{R}^f) 
    &= \sum_i \sum_j a_{ij} ||\mat{r}^e_i - \mat{r}^f_j ||^2 \\
    &= (\mat{R}^e - \mat{R}^f)^\top \mat{A} (\mat{R}^e - \mat{R}^f).
\end{align}
where subscript $\mat{A}$ indicates that the alignments are fixed (given). $\mat{A}$ captures the 
relationships between all $V^e$ words in English
with respect to all $V^f$ words in French, and is indeed also
the source of the two main challenges in this formulation, namely: 
\begin{enumerate}
  \item how to derive or learn which words to pair as translation pairs (i.e.\ 
    deriving/learning $\mat{A}$); 
  \item how to \emph{efficiently} evaluate $\Omega(\cdot)$ during training, since 
    naively evaluating it scales as the product of the two vocabulary sizes 
    $O(|V^e| \cdot |V^f|)$ \emph{at each training step}.
\end{enumerate}

The cross-lingual objective therefore penalizes the Euclidian distance between
words in the two embedding spaces ($\mathbf{R}^e$ and $\mathbf{R}^f$)
proportional to their alignment frequency. Previous work approached this step
by performing a word-alignment step prior to training to learn the alignment
matrix $\mat{A}$.  However, performing word alignment requires running
Giza\plpl~\cite{och03giza} or FastAlign~\cite{dyer2013simple} software and
training HMM word-alignment models. This is both computationally costly and
also noisy. We would like to learn the translation correspondences without
utilizing word alignments.  In order to do that, we directly exploit the
parallel training data.  The main contribution of this work is to approximate
the costly $\Omega(\cdot)$ term, defined in equation \ref{eqn:xling-full} in terms of
the \emph{global} word-\emph{alignment} statistics, using cheaply-obtained 
\emph{local} word \emph{co-occurrence} statistics obtained from raw-text parallel sentence-pairs
(i.e.\ \emph{without} running a word-alignment step).  The main concept is illustrated
schematically in Figure~\ref{fig:bilbowa-approximation}, and discussed in more
detail below.

As a first step,
notice that since the alignment weights can be normalized to sum to one, we can interpret the
alignment weights as a distribution and write equation \ref{eqn:xling-full} as an
expectation over the distribution of English and French word-alignment
probabilities $a_{ij} = P(w^e_i, w^f_j)$,
\begin{equation}  \label{eqn:bilbowa-xling-expanded}
  \Omega_\mat{A}(\mat{R}^e, \mat{R}^f) 
   = \E_{(i,j) \sim P(w^e, w^f)} \left [ ||\mat{r}^e_i - \mat{r}^f_j ||^2 \right ]
\end{equation}
Since the parallel data is paired at the sentence level, we know that translation pairs for
the \emph{en} sentence occur in the \emph{fr} sentence, but we do not know
where. We therefore need a word-alignment model. 
A naive assumption is to assume that each observed
\emph{en} word can potentially be aligned with each observed \emph{fr} word (i.e.\
to assume a \bld{uniform word-alignment model}), for each word \emph{in the
observed sentence pairs}.  Under this assumption, one can then approximate
equation \ref{eqn:bilbowa-xling-expanded} by sampling $S$ $m$-length English and
$n$-length French sentence-pairs $(s^{e},s^{f})$ from the parallel training
data: 
\begin{equation} \label{eqn:bilbowa-xling-approx}
  \Omega_\mat{A}(\mat{R}^e, \mat{R}^f) 
    \approx \frac{1}{S} \sum_{(s^{e},s^{f}) \in S} 
       \frac{1}{mn} \sum_{w_i \in s^{e}} \sum_{w_j \in s^{f}} 
         ||\mat{r}^e_i - \mat{r}^f_j ||^2
\end{equation}
It is important to note that the lengths of the sampled English and French parallel 
sentences $m$ and $n$ need not be equal, and more importantly $m \ll |V^e|$ and 
$n \ll |V^f|$. Furthermore, notice that under a uniform alignment model, at each 
training step, each word
in the sampled English sentence $s^e$ will be updated towards all words in the
French sentence $s^f$. We can precompute this by simply updating each English
word towards the mean French bag-of-words \bld{sentence-vector}. Specifically,
we implement equation \ref{eqn:bilbowa-xling-approx} by sampling only one parallel 
sentence-pair per training step (i.e.\ $S=1$), and at each
training step $t$ we optimize the following \bld{sampled, approximate cross-lingual
objective}:
\begin{equation}
    \label{eqn:bilbowa-xling-sentvec}
    \Omega^{(t)}_\mat{A}(\mat{R}^e, \mat{R}^f) \triangleq 
    || \frac{1}{m} \sum_{w_i \in s^e}^m \mat{r}^e_i - 
    \frac{1}{n} \sum_{w_j \in s^f}^n \mat{r}^f_j ||^2
\end{equation}
where $s^*$ denotes the English or French sampled sentence-pair drawn from the
parallel corpus.  In other words, \emph{the BilBOWA-loss minimizes a sampled
$L_2$-loss between the mean bag-of-words sentence-vectors of the parallel corpus}.
On its own, this objective is degenerate since all embeddings would converge to
the trivial solution (by collapsing all embeddings to the same value), but
coupled as a regularizer with the monolingual losses, we find that it works
very well in practice. By sampling training sentences from the parallel document
distribution, this objective efficiently approximates equation~\ref{eqn:xling-full} 
(the more two words are observed together in a parallel sentence-pair, the
stronger the embeddings for the two words will be pushed together, i.e.\
proportional to $a_{ij}$), without having to actually compute the word 
alignment weights $a_{ij}$.

\subsection{Parallel subsampling for better results}
Equation \ref{eqn:bilbowa-xling-sentvec} is an approximation of
equation~\ref{eqn:xling-full}.  As illustrated in
Figure~\ref{fig:bilbowa-approximation}, we are really interested in estimating
the global word-alignment statistics for a word-pair, i.e.\ $a_{ij}$.
However, by sampling words at the sentence-level, the local alignment
statistics are skewed by the words' \emph{unigram frequencies of occurrence} in
a given sentence (i.e.\ regardless of alignment). Since language has a very
strong Zipfian distribution we therefore find in practice that
equation~\ref{eqn:bilbowa-xling-sentvec} \emph{over-regularizes the frequent
words}.  A simple solution to this is to subsample (discard) words from the
parallel sentences proportional to their unigram probabilities of occurrence.
In other words, we discard each word from the parallel sentences with a probability 
that depends on its unigram frequency of occurrence. This heavily discards frequent words
and effectively flattens the unigram distribution to a uniform distribution. This idea
is closely related to the monolingual subsampling employed in the word2vec models,
although it is motivated for a different reason in the cross-lingual setting to obtain 
a better approximation of the global word-alignment statistics from the local sentence-level 
co-occurrence statistics (see Figure~\ref{fig:bilbowa-approximation}).

\begin{figure}
  \centering
  \includegraphics[width=0.45\textwidth]{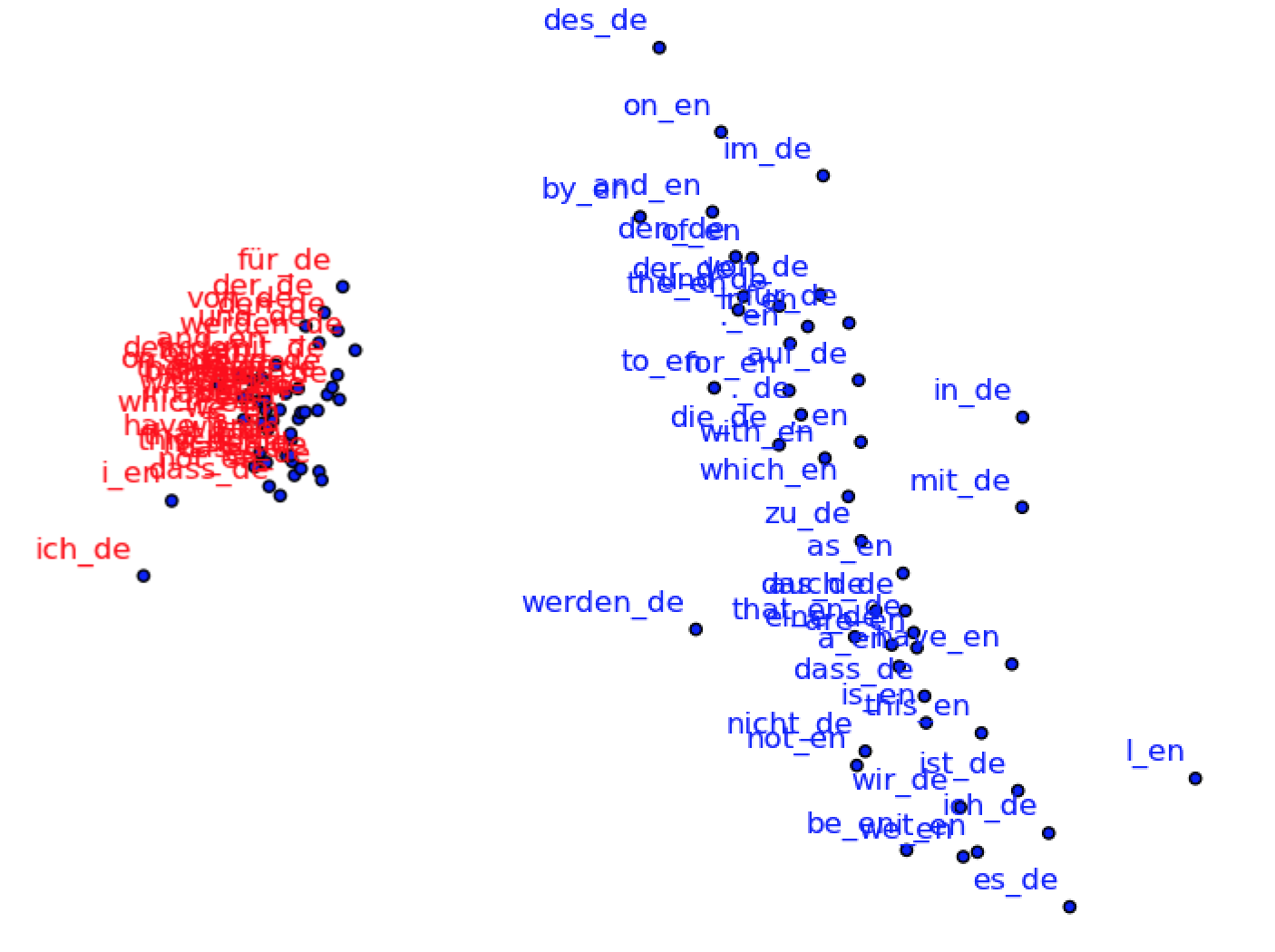} 
    \caption{A joint t-SNE visualization of the \emph{same} 25 most frequent English and German
  words trained \emph{without} (red, left) and \emph{with} parallel subsampling (blue, right),
  illustrating the effect that occurs without parallel subsampling,
  as frequent words are over-regularized towards the origin.}
   \label{fig:subsample}
\end{figure}

In practice we found this useful to learn finer-grained cross-lingual embeddings
for the frequent words. To better illustrate the effect this has on training,
we \emph{jointly} visualized 
the top-25
most frequent words in English and German using the t-SNE algorithm. This is
illustrated in Figure~\ref{fig:subsample}.  We show in red the embeddings
trained without subsampling and in blue the embeddings for the same words trained 
using parallel subsampling. As the visualization shows, without subsampling frequent words are
over-regularized and cluster near the origin. This effect is largely reduced by
the proposed subsampling scheme.

\section{Implementation and Training Details}
We implemented our model in C by building on the popular open-source
\texttt{word2vec} toolkit\footnote{\url{https://code.google.com/p/word2vec/}}.
The implementation launches a monolingual skipgram model as a separate thread
for each language, as well as a cross-lingual thread. All threads access
the shared embedding parameters asynchronously. For training the model, we make
use of online asynchronous stochastic gradient descent (ASGD), where
at time step $t$, parameter $\theta$ is updated as 
\begin{equation}
    \label{eqn:bilbowa-sgd}
    \theta^{(t)} = \theta^{(t-1)} - \eta \frac{\partial \loss}{\partial \theta}
\end{equation}
Our initial implementation synchronized updates between threads, but we found
that simply clipping individual updates to $[-0.1,0.1]$ per thread was sufficient to
ensure training stability and considerably improved training speed.  For
monolingual training data, we use the freely available, pre-tokenized
Wikipedia datasets~\cite{polyglot2013}. For cross-lingual training we use the
freely-available Europarl v7 corpus~\cite{koehn2005europarl}. Unlike
the approach of \cite{klementiev2012} however, we do not need to perform a word-alignment step
first. Instead our implementation trains directly on the raw parallel text
files obtained after applying the standard preprocessing scripts that come with
the data to tokenize, recase and remove all empty sentence-pairs. Embedding
matrices were initialized by drawing from a zero mean, unit-variance gaussian
distribution. The skipgram negative sampling objectives (a simplified form
of noise-contrastive estimation) require us to sample $k$ noise words per training 
pair from the unigram $P(w)$ \emph{en} and \emph{fr}  distributions, and we 
set $k=15$ which has been shown to give good results.


Doing each training update therefore occurs asynchronously across the threads.
Monolingual threads each select a context-target ($h$,$w_t$)-pair for each language 
and sample $k$ noise words according to their unigram noise distributions.  The cross-lingual
thread samples a random pair of parallel sentences from the parallel data. Finally, 
each thread makes an update to all parameters asynchronously according to equation 
\ref{eqn:bilbowa-sgd}, for which gradients are easy to compute due to the square-loss
of the BilBOWA-loss and the log-linear nature of the skipgram models.  The learning 
rate was set to 0.1, with linear decay.

\section{Experiments}
In this section we present experiments  which evaluate the utility of the
induced representations.  We evaluate the embeddings in a cross-lingual document 
classification task which tests \emph{semantic transfer} of information across languages,
as well as a word-level translation task which tests fine-grained \emph{lexical transfer}.

\subsection{Cross-lingual Document Classification}
\label{sec:cldc}

We use an \emph{exact replication} of the 
cross-lingual document classification (\textbf{CLDC}) setup introduced by Klementiev et
al.~\cite{klementiev2012} to evaluate their cross-lingual embeddings. The
CLDC task setup is as follows: The goal is to classify documents in a target
language using only labelled documents in a source language. Specifically,
we train an averaged perceptron classifier on the labelled training data in the 
source language and then attempt to apply the classifier as-is to the target data (known as
``direct transfer''). Documents are represented as the tf-idf-weighted
sum of the embedding vectors of the words that appear in the documents.

Similar to Klementiev, we induce cross-lingual
embeddings for the English-German language pair, and use the induced
representations to classify a subset of the English and German sections of
the Reuters  RCV1/RCV2 multilingual corpora \cite{lewis2004rcv1} as pertaining
to one of four categories: \texttt{CCAT} (Corporate/Industrial),
\texttt{ECAT} (Economics), \texttt{GCAT} (Government/Social), and
\texttt{MCAT} (Markets).

For the classification experiments, 15,000 documents (for each language) were
randomly selected from the RCV1/2 corpus, with one third (5,000) used as the
test set and the remainder divided into training sets of
sizes between 100 and 10,000, and a separate, held-out validation set
of 1,000 documents used during the development of our models.
Since our setup exactly mirrors Klementiev et al, we use the same baselines,
namely: the \emph{majority} class baseline, \emph{glossed} (replacing words in
the target document by their most frequently aligned words in the source
language), and a stronger \emph{MT} baseline (translating target documents into
the source language using an SMT system).

\begin{table*}
  \small
    \centering
    \begin{tabular}{l|cc|r}
      \hline
      \bld{Method}  & \emph{en} $\rightarrow$ \emph{de} & \emph{de} $\rightarrow$ \emph{en}  & \bld{Training Time (min)} \\
      \hline
      \emph{Majority Baseline} & 46.8 & 46.8 & - \\
      \emph{Glossed Baseline} & 65.1 & 68.6 & - \\
      \emph{MT Baseline} & 68.1 & 67.4 & - \\
      \hline
      Klementiev et al.\              & 77.6       & 71.1     & 14,400 \\
      Bilingual Auto-encoders (BAEs)  & \bld{91.8} & 72.8     & 4,800  \\
      BiCVM                           & 83.7       & 71.4     & 15     \\
      \bld{BilBOWA (this work)}             & 86.5       & \bld{75} & \bld{6}          \\
      \hline
    \end{tabular}
    \caption{Classification accuracy and training times for the proposed BilBOWA method compared
    to Klementiev et al.~\cite{klementiev2012}, Bilingual Auto-encoders~\cite{chandar2014autoencoder},
    and the BiCVM model~\cite{hermann2013multilingual}, on an exact replica of the
    Reuters cross-lingual document classification task. These methods were all used
    to induce 40-dimensional embeddings using the same training data. Baseline results 
    are from Klementiev.}
    \label{tab:cldc-results}
\end{table*}

Results are summarized in Table~\ref{tab:cldc-results}. In order to make all
results comparable, results for all methods reported here were obtained using
the same embedding dimensionality of $40$ and the same training data.  We use
the first 500K lines of the English-German Europarl data \emph{both} as
monolingual and parallel training data. We use a vocabulary size of $46,678$
for English and $47,903$ for German. Since our method is motivated as a faster
version of the model proposed by Klementiev \emph{et al.}, we note that we 
significantly improve upon their results, while
training in $6$ minutes versus the original 10 days (14,400 minutes). This yields a
total factor $2,400$ speedup. This demonstrates that the BilBOWA loss 
(equation \ref{eqn:bilbowa-xling-sentvec}) is both a
computationally-efficient and an accurate approximation of the full cross-lingual
objective implemented by Klementiev (equation \ref{eqn:xling-full}).

Next, we compare our method to the current state-of-the-art bilingual embedding
methods.  The current state-of-the-art on this task is $91.8$ (en2de) and
$72.8$ (de2en) reported using the Bilingual Auto-encoder (BAE) model by
\cite{chandar2014autoencoder}. Hermann et al.~\cite{hermann2013multilingual}
report $83.7$ and $71.4$ with the BiCVM model. As shown, our model outperforms
the BiCVM on both tasks, and outperforms BAEs on German to English to yield a
current state-of-the-art result on that task of $75$\%. The runtime of our
method also compares very favorably to other methods. Note that even though the
BiCVM method should in principle be as fast or faster than our method, its
reported training time here is slightly higher since it was trained for more iterations
over the data.

\subsection{WMT11 Word Translation}
\label{sec:wmt}

We also evaluated the induced cross-lingual embeddings on the word
translation task used by Mikolov et al.~\cite{mikolov2013exploiting} using the
publicly-available WMT11 data\footnote{\url{http://www.statmt.org/wmt11/}}. 
In this task, the authors extracted the $6K$ most frequent words from the WMT11
English-Spanish data, and then used the online Google Translate service to
derive dictionaries by translating these source words into the target language
(individually for English and Spanish).  Since their method requires
translation-pairs for training, they used the first $5K$ most frequent words to
learn the ``translation matrix'', and then evaluated their method on the
remaining $1K$ words used as a test set. To translate a source word, one finds
its $k$ nearest neighbours in the target language embedding space, and then
evaluate the translation precision $P@k$ as the fraction of target translations
that are within the top-$k$ words returned using the specific method.  Our
method does not require translation-pairs for training, so we simply test on
the same $1K$ test-pairs.

We use as baselines the same two methods described in ~\cite{mikolov2013exploiting}. 
\emph{Edit Distance} ranks words based on their edit-distance. \emph{Word Co-occurrence} is
based on distributional similarity: For each word $w$, one first constructs a
word co-occurrence vector which counts the words with which $w$ co-occurs
within a 10-word window in the corpus. The word-count vectors are then mapped from the source
to the target language using the dictionary. Finally, for each test word, the word
with the most similar vector in the target language is selected as its translation.

\begin{table*}
  \small
  \centering 
  \begin{tabular}{l|c|c||c|c}
    \hline
    \bld{Method} & \bld{En$\rightarrow$Sp P@1} &  \bld{Sp$\rightarrow$En P@1} &  \bld{En$\rightarrow$Sp P@5} & \bld{Sp$\rightarrow$En P@5} \\
    \hline
    Edit Distance & 13 & 18 & 24 & 27 \\
    Word Co-occurrence & 30 & 19 & 20 & 30 \\
   \emph{Mikolov et al.}, 2013 & 33 & 35 & \bld{51} & 52 \\
   \textbf{BilBOWA (this work)} & \bld{39} (+6) & \bld{44} (+9) & \bld{51} & \bld{55} (+3) \\
   \hline
  \end{tabular}
  \caption{Results for the translation task measured as word translation
  accuracy (out of 100, higher is better) evaluated on the top-1 and top-5 words 
  as ranked by the method. Cross-lingual embeddings are induced and distance in 
  the embedded space are used to select word translation pairs. $+x$ indicates 
  improvement in absolute precision over the previous state-of-the-art on this 
  task~\cite{mikolov2013exploiting}. }
  \label{tab:wmt11-task}
\end{table*}

The results on the English-Spanish translation tasks are summarized in
Table~\ref{tab:wmt11-task}. We induced 40-dimensional embeddings using the
English and Spanish Wikipedias and Europarl as parallel data. Our model
improves on both the baselines and on \emph{Mikolov et al.}'s method on both
tasks and gives a noticeable improvement in accuracy for the $P@1$ prediction.
For the English to Spanish translation, we improve \emph{absolute word
translation accuracy} by 6 percent. For the Spanish to English task, we improve
absolute word translation accuracy by 9 percent. This indicates that our model
is able to learn fine-grained translation equivalences from the monolingual
data by using only the raw-text, sentence-aligned parallel data, despite the
lack of word-level alignments or training dictionaries.

\section{Discussion}

The BilBOWA model as introduced in this paper utilizes a sampled $L_2$
bag-of-words cross-lingual loss. This is the main source of the significant
speedup over the Klementiev model, allowing training to scale to much larger
datasets, which in turn yields more accurate features.  We found that the
asynchronous implementation significantly speeds up training with no noticeable 
impact on the quality of the learned embeddings, and the parallel
subsampling improves the accuracy of the learned features.  Getting
asynchronous training to work required clipping the updates, especially as the
dimensionality of the embeddings gets larger. Parallel subsampling makes training 
more accurate, especially for the frequent
words, and therefore turns out to be important both in the monolingual and
crosslingual setting. We have motivated the reason for the crosslingual setting
as helping to uncover a better approximation of the global alignment statistics
from the observed local, sentence-level co-occurrences.

Despite the speedup, the model is still much slower to use than offline methods
like translation matrix~\cite{mikolov2013exploiting} or multilingual
CCA~\cite{faruqui2014improving}.  However, results on the translation task
suggest BilBOWA can learn finer-grained cross-lingual relationships than these
methods, and can train over much larger monolingual datasets than parallel-only
methods. If the goal is to learn high-quality general purpose bilingual
embeddings, it is always beneficial to leverage more training data, and
hence a hybrid model like BilBOWA might be a better choice than a parallel-only
technique like BiCVM or BAEs.

\section{Conclusion}
We introduce BilBOWA, a computationally-efficient model for inducing bilingual
distributed word representations directly from monolingual raw text and a limited
amount of parallel data, without requiring word-alignments or dictionaries.  
BilBOWA combines advances in training monolingual word embeddings with a
particularly efficient novel sampled cross-lingual objective.  The result is
that the required computations per training step scales only with the number of
words in the sentences, thereby enabling efficient large-scale cross-lingual
training.  We achieve state-of-the-art results for English-German cross-lingual
document classification whilst obtaining up to three orders of magnitude speedup, and 
improve upon the previous state of the art in an English-Spanish word-translation task.



\small
\bibliography{fast-biling}

\begin{thebibliography}{23}
\providecommand{\natexlab}[1]{#1}
\providecommand{\url}[1]{\texttt{#1}}
\expandafter\ifx\csname urlstyle\endcsname\relax
  \providecommand{\doi}[1]{doi: #1}\else
  \providecommand{\doi}{doi: \begingroup \urlstyle{rm}\Url}\fi

\bibitem[Al-Rfou' et~al.(2013)Al-Rfou', Perozzi, and Skiena]{polyglot2013}
Al-Rfou', Rami, Perozzi, Bryan, and Skiena, Steven.
\newblock Polyglot: Distributed word representations for multilingual nlp.
\newblock In \emph{Proceedings of the Seventeenth Conference on Computational
  Natural Language Learning}, pp.\  183--192, Sofia, Bulgaria, August 2013.
  Association for Computational Linguistics.
\newblock URL \url{http://www.aclweb.org/anthology/W13-3520}.

\bibitem[Bengio et~al.(2003)Bengio, Ducharme, and Vincent]{bengio2003neural}
Bengio, Y, Ducharme, R, and Vincent, P.
\newblock A neural probabilistic language model.
\newblock \emph{Journal of Machine Learning Research}, 2003.
\newblock URL \url{http://ukpmc.ac.uk/abstract/CIT/412956}.

\bibitem[Bengio \& Senecal(2008)Bengio and Senecal]{bengio2008adaptive}
Bengio, Yoshua and Senecal, J-S.
\newblock Adaptive importance sampling to accelerate training of a neural
  probabilistic language model.
\newblock \emph{Neural Networks, IEEE Transactions on}, 19\penalty0
  (4):\penalty0 713--722, 2008.

\bibitem[Blitzer et~al.(2006)Blitzer, McDonald, and Pereira]{blitzer2006domain}
Blitzer, J., McDonald, R., and Pereira, F.
\newblock Domain adaptation with structural correspondence learning.
\newblock In \emph{Conference on Empirical Methods in Natural Language
  Processing}, Sydney, Australia, 2006.

\bibitem[Chandar et~al.(2014)Chandar, Lauly, Larochelle, Khapra, Ravidran,
  Raykar, and Saha]{chandar2014autoencoder}
Chandar, Sarath, Lauly, Stanislas, Larochelle, Hugo, Khapra, Mitesh~M.,
  Ravidran, Balaraman, Raykar, Vikas, and Saha, Amrita.
\newblock An autoencoder approach to learning bilingual word representations.
\newblock \emph{Proceedings of NIPS 2014}, 2014.

\bibitem[Collobert et~al.(2011)Collobert, Weston, Bottou, Karlen, Kavukcuoglu,
  and Kuksa]{collobert2011natural}
Collobert, R., Weston, J., Bottou, L., Karlen, M., Kavukcuoglu, K., and Kuksa,
  P.
\newblock Natural language processing (almost) from scratch.
\newblock \emph{Journal of Machine Learning Research}, 12:\penalty0 2493--2537,
  2011.

\bibitem[Daum{\'e}~III(2009)]{daume2009frustratingly}
Daum{\'e}~III, Hal.
\newblock Frustratingly easy domain adaptation.
\newblock In \emph{Proceedings of the 48th Annual Meeting of the Association
  for Computational Linguistics}. Association for Computational Linguistics,
  2009.

\bibitem[Dyer et~al.(2013)Dyer, Chahuneau, and Smith]{dyer2013simple}
Dyer, Chris, Chahuneau, Victor, and Smith, Noah~A.
\newblock A simple, fast, and effective reparameterization of ibm model 2.
\newblock ACL, 2013.

\bibitem[Faruqui \& Dyer(2014)Faruqui and Dyer]{faruqui2014improving}
Faruqui, Manaal and Dyer, Chris.
\newblock Improving vector space word representations using multilingual
  correlation.
\newblock In \emph{Proceedings of EACL 2014}, 2014.

\bibitem[Goldberg \& Levy(2014)Goldberg and Levy]{goldberg2014word2vec}
Goldberg, Yoav and Levy, Omer.
\newblock word2vec explained: Deriving mikolov et al.Õs negative-sampling
  word-embedding method.
\newblock \emph{arXiv preprint arXiv:1402.3722}, 2014.

\bibitem[Hermann \& Blunsom(2013)Hermann and Blunsom]{hermann2013multilingual}
Hermann, Karl~Moritz and Blunsom, Phil.
\newblock Multilingual distributed representations without word alignment.
\newblock \emph{arXiv preprint arXiv:1312.6173}, 2013.

\bibitem[Klementiev et~al.(2012)Klementiev, Titov, and
  Bhattarai]{klementiev2012}
Klementiev, Alexandre, Titov, Ivan, and Bhattarai, Binod.
\newblock Inducing crosslingual distributed representations of words.
\newblock In \emph{Proceedings of the International Conference on Computational
  Linguistics (COLING)}, Bombay, India, December 2012.

\bibitem[Koehn(2005)]{koehn2005europarl}
Koehn, Philipp.
\newblock Europarl: A parallel corpus for statistical machine translation.
\newblock In \emph{MT summit}, volume~5, pp.\  79--86, 2005.

\bibitem[Lewis et~al.(2004)Lewis, Yang, Rose, and Li]{lewis2004rcv1}
Lewis, David~D, Yang, Yiming, Rose, Tony~G, and Li, Fan.
\newblock Rcv1: A new benchmark collection for text categorization research.
\newblock \emph{The Journal of Machine Learning Research}, 5:\penalty0
  361--397, 2004.

\bibitem[Mikolov et~al.(2013{\natexlab{a}})Mikolov, Le, and
  Sutskever]{mikolov2013exploiting}
Mikolov, Tomas, Le, Quoc~V, and Sutskever, Ilya.
\newblock Exploiting similarities among languages for machine translation.
\newblock In \emph{International Conference on Learning Representations
  (ICLR)}, 2013{\natexlab{a}}.

\bibitem[Mikolov et~al.(2013{\natexlab{b}})Mikolov, Sutskever, Chen, Corrado,
  and Dean]{mikolov2013distributed}
Mikolov, Tomas, Sutskever, Ilya, Chen, Kai, Corrado, Greg~S, and Dean, Jeff.
\newblock Distributed representations of words and phrases and their
  compositionality.
\newblock In \emph{Advances in Neural Information Processing Systems}, pp.\
  3111--3119, 2013{\natexlab{b}}.

\bibitem[Mnih \& Teh(2012)Mnih and Teh]{mnih2012fast}
Mnih, Andriy and Teh, Yee~Whye.
\newblock A fast and simple algorithm for training neural probabilistic
  language models.
\newblock In \emph{Proceedings of the 29th International Conference on Machine
  Learning (ICML)}, 2012.

\bibitem[Och \& Ney(2003)Och and Ney]{och03giza}
Och, Franz~Josef and Ney, Hermann.
\newblock A systematic comparison of various statistical alignment models.
\newblock \emph{Computational Linguistics}, 29\penalty0 (1):\penalty0 19--51,
  2003.

\bibitem[Pan \& Yang(2010)Pan and Yang]{pan2010survey}
Pan, Sinno~Jialin and Yang, Qiang.
\newblock A survey on transfer learning.
\newblock \emph{Knowledge and Data Engineering, IEEE Transactions on},
  22\penalty0 (10):\penalty0 1345--1359, 2010.

\bibitem[Pennington et~al.(2014)Pennington, Socher, and
  Manning]{pennington2014glove}
Pennington, Jeffrey, Socher, Richard, and Manning, Christopher~D.
\newblock Glove: Global vectors for word representation.
\newblock \emph{Proceedings of the Empiricial Methods in Natural Language
  Processing (EMNLP 2014)}, 12, 2014.

\bibitem[Turian et~al.(2010)Turian, Ratinov, and Bengio]{turian2010word}
Turian, J., Ratinov, L., and Bengio, Y.
\newblock Word representations: A simple and general method for semi-supervised
  learning.
\newblock In \emph{Proceedings of the 48th Annual Meeting of the Association
  for Computational Linguistics}, pp.\  384--394. Association for Computational
  Linguistics, 2010.

\bibitem[Walker(1977)]{walker1977efficient}
Walker, Alastair~J.
\newblock An efficient method for generating discrete random variables with
  general distributions.
\newblock \emph{ACM Transactions on Mathematical Software (TOMS)}, 3\penalty0
  (3):\penalty0 253--256, 1977.

\bibitem[Zou et~al.(2013)Zou, Socher, Cer, and Manning]{zoubilingual}
Zou, Will~Y, Socher, Richard, Cer, Daniel, and Manning, Christopher~D.
\newblock Bilingual word embeddings for phrase-based machine translation.
\newblock In \emph{Proceedings of the 2013 Conference on Empirical Methods in
  Natural Language Processing (EMNLP)}, 2013.

\end{thebibliography}
\bibliographystyle{icml2015}

\end{document}